\documentclass[lettersize,journal]{IEEEtran}
\usepackage{amsmath,amsfonts}
\usepackage{algorithmic}
\usepackage{array}
\usepackage[caption=false,font=normalsize,labelfont=sf,textfont=sf]{subfig}
\usepackage{textcomp}
\usepackage{stfloats}
\usepackage{url}
\usepackage{verbatim}
\usepackage{graphicx}

\usepackage{hyperref}
\usepackage{mathtools}
\usepackage {amsmath}
\usepackage{amsfonts}
\usepackage{tabularray}

\usepackage[ruled,linesnumbered]{algorithm2e}

\usepackage{lipsum}
\usepackage{multicol}
\usepackage{graphicx}
\usepackage{amssymb}
\usepackage{subfig}
\usepackage{multirow}
\usepackage{float}
\usepackage{balance}
\usepackage{caption}
\usepackage{blindtext}
\usepackage{hhline}

\def\BibTeX{{\rm B\kern-.05em{\sc i\kern-.025em b}\kern-.08em
    T\kern-.1667em\lower.7ex\hbox{E}\kern-.125emX}}

\let\oldtwocolumn\twocolumn
\renewcommand\twocolumn[1][]{%
    \oldtwocolumn[{#1}{
    \begin{center}
           \includegraphics[width=\textwidth]{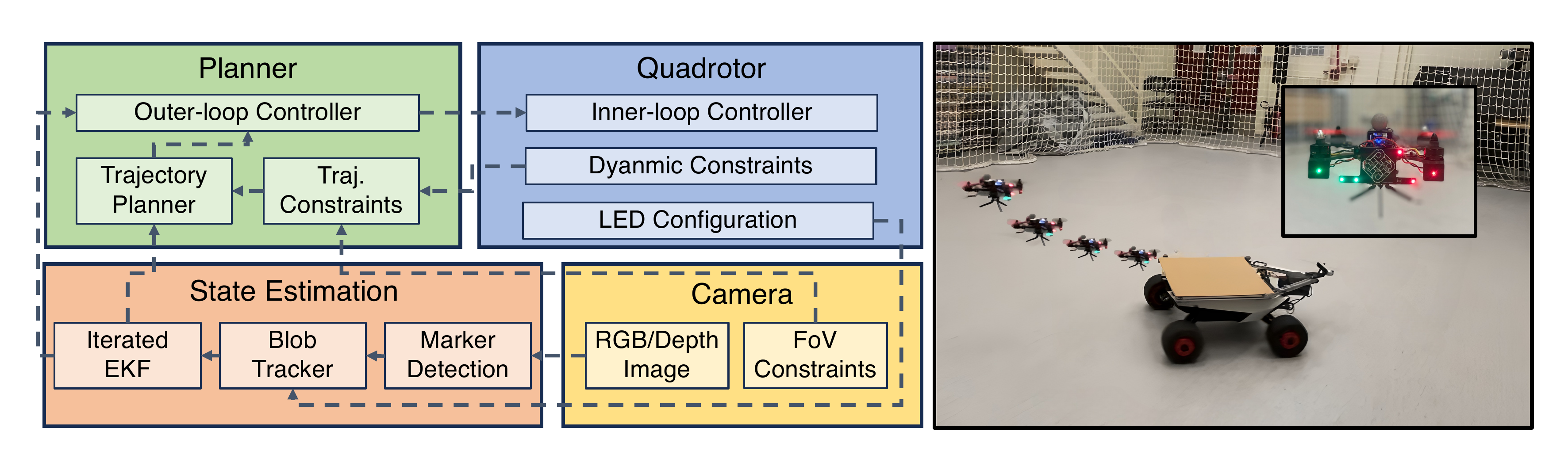}
           \captionof{figure}{Hardware and software details of the proposed non-robocentric landing system. The system could be separately discussed in two main parts: state estimation and planner. Note that all modules above are located on the ground, with no exteroceptive sensors and computing units on the quadrotor. }
           \label{fig:cover}
        \end{center}
    }]
}

\begin{document}
\title{Towards Non-Robocentric Dynamic Landing of Quadrotor UAVs}
\author{
    Li-Yu Lo$^{1,2}$ \IEEEmembership{Student Member, IEEE}, 
    Boyang Li$^{3}$, 
    Chih-Yung Wen$^{1,2}$, 
    and Ching-Wei Chang$^{4\dagger}$

  \thanks{$^{1}$AIRo-LAB, Department of Aeronautical and Aviation Engineering, 
      The Hong Kong Polytechnic University, 
      Kowloon, Hong Kong.
  }
  \thanks{$^{2}$Research Centre for Unmanned Autonomous System, 
      The Hong Kong Polytechnic University, 
      Kowloon, Hong Kong.
  }
  \thanks{
    $^{3}$School of Engineering, The University of Newcastle, 
      Callaghan, NSW 2308, Australia.
	}
  \thanks{
    $^{4}$Hong Kong Center for Construction Robotics,
    The Hong Kong University of Science and Technology, 
	New Territories, Hong Kong.
  }
  \thanks{E-mails: \texttt{\{patty.lo, boyang.li, chihyung.wen, ccw\}
	@\{connect.polyu.hk, newcastle.edu.au, polyu.edu.hk, ust.hk\}
	}
  }
  \thanks{
    $^{\dagger}$ corresponding author
  }%
}

\markboth{PREPRINT}%
{}
\maketitle

\begin{abstract}
	In this work, we propose a dynamic landing solution without the need for onboard exteroceptive sensors and an expensive computation unit, where all localization and control modules are carried out on the ground in a non-inertial frame. Our system starts with a relative state estimator of the aerial robot from the perspective of the landing platform, where the state tracking of the UAV is done through a set of onboard LED markers and an on-ground camera; the state is expressed geometrically on manifold, and is returned by Iterated Extended Kalman filter (IEKF) algorithm. Subsequently, a motion planning module is developed to guide the landing process, formulating it as a minimum jerk trajectory by applying the differential flatness property. Considering visibility and dynamic constraints, the problem is solved using quadratic programming, and the final motion primitive is expressed through piecewise polynomials. Through a series of experiments, the applicability of this approach is validated by successfully landing 18 cm \texttimes \space 18 cm quadrotor on a 43 cm \texttimes \space 43 cm platform, exhibiting performance comparable to conventional methods. Finally, we provide comprehensive hardware and software details to the research community for future reference.
\end{abstract}

\begin{IEEEkeywords}
Quadrotors, Dynamic Landing, Relative Pose Estimation, Motion Planning.
\end{IEEEkeywords}

\section*{Supplementary Material}
\label{sec:SM}
Supporting video of this paper is available at \url{https://youtu.be/7wiCh46MQmc}. Hardware and software are detailed at: \url{https://github.com/HKPolyU-UAV/alan.git}

\section{Introduction}
\label{sec:intro}
\IEEEPARstart{D}{ue} 
    to its high versatility and easy deployment, unmanned aerial systems, especially multirotor UAVs like quadrotors, have garnered significant attention in academia and industry over the past decade \cite{tokekar2016sensor,sun2016camera}. Safe takeoff and landing capabilities are crucial for the success of aerial missions, particularly in scenarios involving frequent releasing and retracting of drones, as seen in large-scale deployments for tasks such as marine surveying \cite{zhou2022development} or drone-assisted parcel delivery \cite{dissanayaka2023review}. The dynamic performance of landing tasks is often essential for efficiency despite introducing challenges and potential threats to mission reliability, making it a topic worthy of investigation.

	However, rather than adopting an on-board approach, we relocate most sensing and computing modules to the ground. Consequently, the airborne quadrotor is equipped solely with a low-cost flight controller unit, which we classify as a ``non-robocentric approach'' in this study, where the aerial robot has no independent sensing and decision-making ability. The argument put forth is that this particular configuration has the potential to significantly reduce the payload of the UAV and simplify its avionics system, while maintaining a performance level comparable to that of a traditional autonomous landing system. Moreover, in cases where a UAV mission does not necessitate the utilization of computationally intensive perception modules (as exemplified by \cite{zhou2022development}), it is posited that the devised system could serve as a more efficient solution, low power-consuming for retracting UAVs. 
	
	Via formulating a non-inertial control problem, the proposed framework relies on state-of-the-art machine vision algorithms, quadrotor motion planning, and control, all of which are executed by the sensing and computing modules situated on the ground. This approach ensures optimal performance while alleviating the computational burden on the airborne quadrotor. Aligned with our previous investigation in \cite{chang2022proactive}, this research endeavor approaches the problem with a refined problem formulation and introduces each submodule with finer details. Contributions are summarized as follows:
	\begin{itemize}
		\item {
			A non-robocentric (offboard sensing and computing) autonomous landing system is presented, along with its problem formulation.		
		}
		\item {
			A relative state estimation module based on an Iterated Extended Kalman filter (IEKF) is proposed for aggressive maneuver in the non-inertial control framework.
		}
		\item {
			By considering the landing safety corridors and dynamic constraints, a method to generate touchdown trajectory is designed to ensure the safety and smoothness of the landing process.	
		}
		\item {
			We integrate all the above-described modules and validate the performance via experiments. All hardware and software implementation details are released for future research. 
		}

	\end{itemize}

    The hardware and software architecture is illustrated in Figure \ref{fig:cover}, where relations between each submodule are shown. The rest of this paper is arranged as follows. Section \ref{sec:related} first mentions some significant pioneering work. Section \ref{sec:problem} then describes the problem definition and the overall software architecture. Section \ref{sec:RSE} further illustrates the adopted method for relative state estimation. In Section \ref{sec:motion}, the motion planning for landing is elucidated. Then, Section \ref{sec:results} will attempt to evaluate the designed system through practical experiments. Lastly, the conclusion is drawn in \ref{sec:conclusion}.
		
\section{Related Work}
\label{sec:related}
	In recent years, autonomous landing for unmanned aerial systems has received significant attention. Simultaneously, the emergence of low-cost computation components has presented an opportunity for the field of machine vision to expand. The aforementioned developments have, therefore, led to a vast body of literature that addresses both static and dynamic landing scenarios. Among these works, Lee et al. \cite{lee2012autonomous} presented an early and prominent study that employed a conventional visual servoing approach to facilitate the landing of a quadrotor on a moving platform. Later on, other visual servoing-based methods for autonomous landing could also be seen in \cite{coutard2011automatic,serra2016landing,wynn2019visual,chen2021auto}.
	Adopting an autonomous navigation framework, Falanga et al. \cite{falanga2017vision} developed a comprehensive vision system for quadcopters by combining vision odometry \cite{forster2016svo} and a trajectory planning module while tracking the relative state of the landing pad with computer vision markers.
	With a slight difference in control approaches, authors in \cite{baca2019autonomous} also presented a UAV system that could land on a real-life ground vehicle through trajectory generation and model predictive control, where GPS, camera, and rangefinder were used as localization and sensing modules. A similar configuration could also be seen in \cite{yang2013onboard,vlantis2015quadrotor,meng2019visual,qi2019autonomous,paris2020dynamic}, where the exploitation of similar vision configuration and control methods was used to address the landing problem.

	Despite the effectiveness of the above work, one could observe that, in many of the discussed scenarios, other than the final landing, the overall flight missions do not necessarily require the installation of vision sensors and high-level embedded computers. For instance, in parcel delivery mentioned in \cite{paris2020dynamic}, an accurate GPS receiver and pre-known terrain information would have sufficed for the mission objectives. From an engineering perspective, although intriguing and effective outcomes have been achieved, solving the autonomous landing problem with onboard sensing might not be the optimal solution on some occasions.

	Compared to onboard sensing and computing methods, a relatively limited body of literature explores ground-based guidance, primarily attributed to its increased complexity in calculating the corresponding pose. A survey conducted by [19] further pointed out that despite such a system allowing more computation resources, the field of view (FoV) is narrower; in addition, mutual communication between ground and UAV is a mandatory requirement. Nevertheless, it is asserted that the former could be resolved by designing the appropriate algorithms while adopting suitable sensors; as for communication modules, all unmanned vehicles are believed to be equipped with primary communication modules such as radio frequency receivers. 
	
	The study presented in \cite{martinez2009trinocular} developed a trinocular guidance system designed explicitly for helicopter landings. The system could acquire accurate 3D translations of the aircraft by employing reconstruction techniques. Building upon this concept, researchers in \cite{kong2013autonomous} and \cite{kong2014ground} proposed a pan-tilt stereo camera unit capable of detecting and tracking UAVs. To enhance robustness in adverse weather conditions, vision systems that work in the infrared spectrum were utilized. Despite some promising results retrieved from the above-listed work, much of the work only dealt with translational data, which is insufficient to control a multirotor fully in the proposed non-robocentric framework. Additionally, only the static landing problem is focused. Contrarily, Ferreira et al. \cite{ferreira20216d} introduced a landing guidance system tailored for dynamic situations. This system utilized deep learning-based methods to estimate the relative 6 degrees of freedom (DoF) pose between a fixed-wing UAV and an aircraft carrier. However, the authors only validated the system through simulation, thus necessitating further practical experiments to ascertain its real-life applicability.
			
	Although effective solutions were seen from the above pioneer works, most literature dealt with the static landing problem. At the same time, a small portion focused on dynamic cases; physical experiments were still lacking. Hence, this research aims to demonstrate a complete ground-based landing guidance system for dynamic cases in which real-life experiments will be conducted. 

\section{Problem Definition}
\label{sec:problem}
    Given a quadrotor with solely an embedded low-cost flight controller and illuminative markers, the offboard sensing and computing devices in the non-inertial frame should determine the relative state information while generating a guidance trajectory to control and land the quadrotor in a non-robocentric fashion. The described problem could be further expressed as follows.

    We first define the relative states of the quadrotor in a local reference frame, where the relative state is defined on the manifold $\mathcal{M}$:
	\begin{align}
		\label{eqn:states}
		\mathbf{x} &= [\mathbf{p},\mathbf{R},\mathbf{v}]^{T} \in \mathcal{M}, \nonumber\\
		where \; \; \mathcal{M} &= \mathbb{R}^{3} \times \mathbb{SO}(3) \times \mathbb{R}^{3}.
	\end{align}

\begin{figure}[t]
		\centering
		\includegraphics[width=0.4\textwidth]{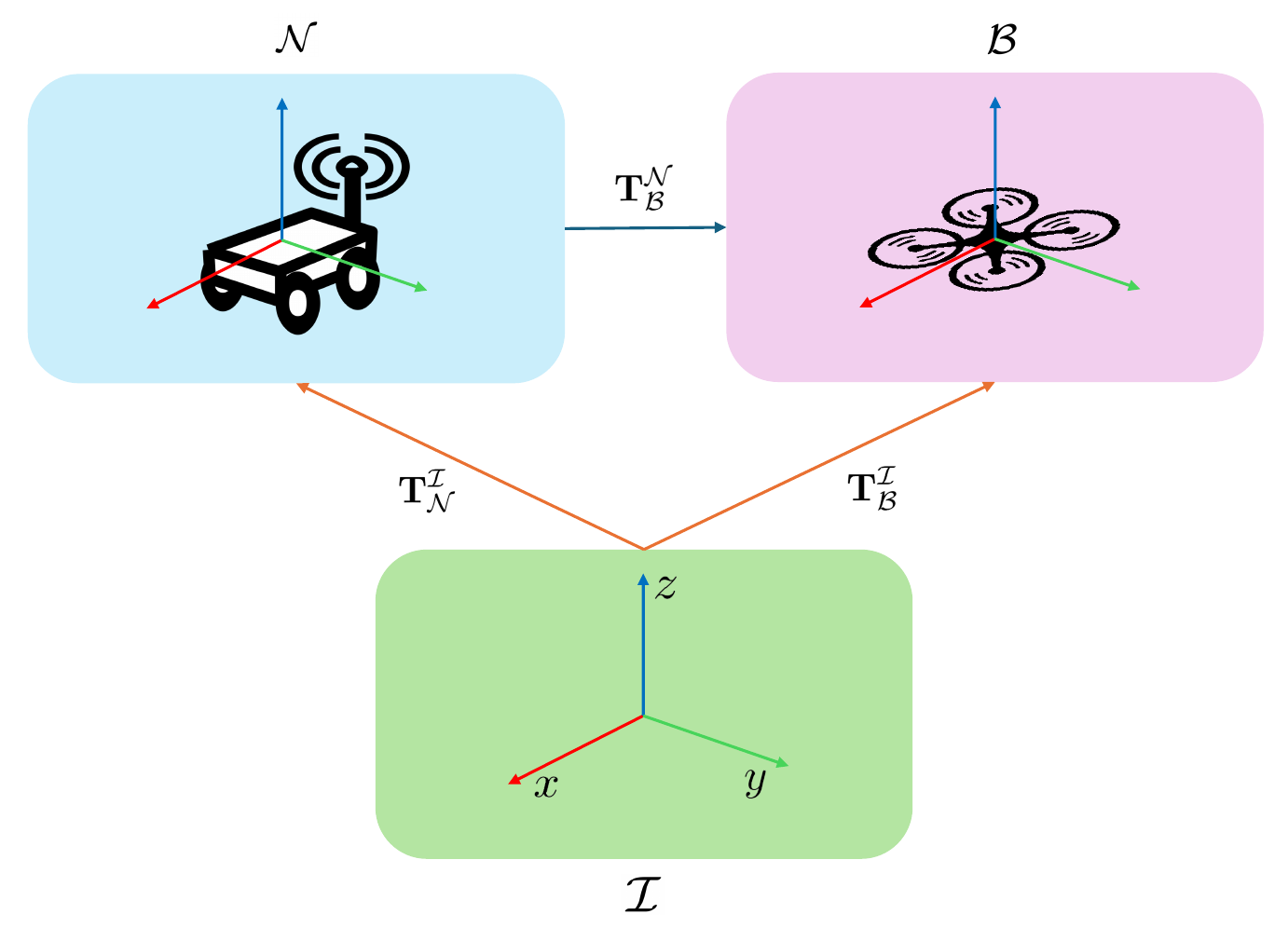}
		\caption{Transformation among inertial frame $\mathcal{N}$, non-inertial local target frame $\mathcal{N}$, and quadrotor body frame $\mathcal{B}$.}
		\label{fig:constellation}
		\vspace{-0.0cm}	 
	\end{figure}

	\noindent In which, $\mathbf{p}$, $\mathbf{R}$, and $\mathbf{v}$ respectively denote the translation, rotation, and linear velocity in the non-inertial frame, denoted as $\mathcal{N}$; $\mathcal{M}$ is the Cartesian product of multiple manifolds. The reason we involve manifold representation is to have easier calculus handling during filtering or optimization \cite{sola2018micro}. The state $\mathbf{x}$ should be subscripted with $k$ to denote the estimation at time step $k$, i.e., $\mathbf{x}_k$. From eqn. \eqref{eqn:states} and figure \ref{fig:constellation}, $\mathbf{p}$ and $\mathbf{R}$ also form $\mathbf{T_{\mathcal{B}}^{\mathcal{N}}}$ belonging to $\mathbb{SE}(3)$, a transformation matrix from the UAV body frame $\mathcal{B}$ to the local non-inertial frame $\mathcal{N}$. We can further derive the UAV pose in the inertial frame $\mathcal{I}$ with UGV poses:
    \begin{align}
      \label{eqn:SO_pose}
	  \mathbf{T_{\mathcal{B}}^{\mathcal{N}}} = \mathbf{T_{\mathcal{I}}^{\mathcal{N}}} \mathbf{T_{\mathcal{B}}^{\mathcal{I}}}. 
    \end{align}

    Meanwhile, we let $\boldsymbol{\sigma}(t)$, parameterized by $t$, be the desired landing trajectory for the quadrotor to track. $\boldsymbol{\sigma}(t)$ comprises translational variable $\mathbf{r}(t)$ and its derivatives, as well as rotational variable $\theta(t)$. 
    \begin{align}
      \label{eqn:SO_traj}
      \boldsymbol{\sigma}(t) = (\mathbf{r}(t), \dot{\mathbf{r}(t)}, \ddot{\mathbf{r}(t)}, \theta(t)),  
      \forall t \in [T_0,T].
    \end{align}	

    \noindent $T_0$ and $T$ in eqn. \eqref{eqn:SO_traj} denotes the starting and terminating time landing trajectory. $\boldsymbol{\sigma(t)}$ should also be subject to constraints for kinematic and dynamic feasibility,
    \begin{align}
      \label{eqn:SO_constr}
      \Lambda_{i} \boldsymbol{\sigma}(t) \leq \boldsymbol{\beta}_{i}, \forall i \in [1, 2,...N].
    \end{align}	

	\noindent In eqn. \eqref{eqn:SO_constr}, $\Lambda_{i}$ is the affine mapping matrix from $\boldsymbol{\sigma}(t)$ to constraints $\boldsymbol{\beta}_i$, where $i$ stands for the number of constraints. The rest of the paper addresses the problem of finding $\mathbf{x}_k$ for $k \in [T^{-},T]$, $T^{-} < T_0$, while generating a constrained $\boldsymbol{\sigma}(t)$ as defined in eqn. \eqref{eqn:SO_traj} and eqn. \eqref{eqn:SO_constr} over $[T_0, T]$. All variables and constraints here are defined in $\mathcal{N}$. The proposed system then adopts the classic closed-loop control structure, where the plant receives inputs based on reference (eqn. \eqref{eqn:SO_traj}) and estimation feedback (eqn. \eqref{eqn:SO_pose}).

\section{Relative State Estimation}
\label{sec:RSE}
	The non-robocentric landing system starts with a robust relative state estimation to get $\mathbf{x}_k$. Herein, a set of airborne LED markers (as shown in figure \ref{fig:cover}) is attached to the quadrotor while the on-ground camera measures the relative state accordingly. The proposed module is inspired by other vision-based relative pose estimation methods \cite{faessler2014monocular} \cite{breitenmoser2011monocular}, but with improvements in terms of robustness for highly dynamic systems such as dynamic landing in this case. 
	
	\subsection{Marker Detection}
	\label{sec:RSE_MD}
	The detection features, i.e., the markers, are attached directly to the quadrotor UAV, where we set $p$ as the number of markers and denote markers translation in quadrotor body frame ($\mathcal{B}$) as $\mathcal{F} = \{f_{1}, f_{2},...f_{p}\}$. On the other hand, a set of detection candidates on 2D image, denoted as $\mathcal{C} = \{c_{1}, c_{2},...c_{q}\}$, is retrieved. As we will be using the perspective-n-point (PnP) algorithm, $p, q$ should be at least 4, and the correspondence pairs $\langle f_i, c_j  \rangle$ should be determined.

	When each 2D image is received, we conduct the detection of $\mathcal{F}$ using conventional image processing methods. In this implementation, as we are using visible LED probe lights, the first objective would be extracting the target light blobs within a noisy image. First, with depth masking $\tau_{dp}$ and gray scaling $\tau_{gy}$, a thresholding function suppresses the unwanted pixels to zero:
	\begin{align}
		\label{eqn:RSE_thres}
		\boldsymbol{\rho} '(u,v)=\left\{\begin{array}{l} 
			\boldsymbol{\rho} (u,v), \;\; if \; \; 
			\boldsymbol{\rho} (u,v)_{dp} < \tau_{dp} 
			\land \boldsymbol{\rho} (u,v)_{gy} > \tau_{gy}\\
			0, \;\; otherwise
			\end{array}\right.
	\end{align}	

	\noindent $\tau_{dp}$ is the effective landing distance, whereas $\tau_{gy}$ depends on the camera's shutter speed. Consequently, the return image is smoothed with Gaussian smoothing, whereas blobs are afterward returned by blobs detection. The 2D coordinates of extracted blobs are then inserted into candidate set $\mathcal{C}$. 
	
	\subsection{Initialization}
	With $\mathcal{F}$ and $\mathcal{C}$, the correspondences could then be searched. Nevertheless, at $k=0$, without any prior information, one can only get the matching relation via a brute-force method. Furthermore, to ensure no outlier detection at start, a stringent initialization condition is set, where we utilize the LED constellation configuration $\mathcal{F}$, the retrieved candidates $\mathcal{C}$, and depth information.
	
	To reject outliers at initialization, the raw 3D information of $\mathcal{C}$ in the camera frame is first acquired with depth information ($s$) and the intrinsic model ($\mathbf{K}$) of the camera:
	\begin{align}
		\label{eqn:PCD}
		\boldsymbol{\zeta}_{i} = \mathbf{K}^{-1} s_i*[u_i,v_i,1]^T,
		\forall \langle u_i,v_i \rangle  \in \mathcal{C}
	\end{align}

	\noindent All $\boldsymbol{\zeta}_{i}$ then forms set $\mathcal{PCD}$. Successively, we compare the mean average deviation ($\text{MAD}$) of $\mathcal{F}$ and $\mathcal{PCD}$,
	\begin{align}
		\label{eqn:RSE_mad}
		\Delta_{\text{MAD}} = \text{MAD}(\mathcal{F}) - \text{MAD}(\mathcal{PCD}).
	\end{align}

	\noindent If $\Delta_{\text{MAD}}$ is below threshold $\lambda_{\text{MAD}}$, the program continues to the next step. Otherwise, it declines initialization. Additionally, to proceed with initialization, $q=|\mathcal{C}|$ should be equal to the size of $p=|\mathcal{F}|$.

	Thereupon, a brute-force search method is used, where raw poses of each correspondence permutation are calculated; the correspondence with the lowest reprojection error is then selected. We first form permutations of $\mathcal{C}$, which is denoted as $\mathcal{Y}$. 
	Meanwhile, as visible LED lights are utilized, additional information from the HSV (hue, saturation, and value) space can reduce the search size. Then, for each permutation ($\mathcal{Y}_j$), a pose $\mathbf{x}_{j}$ is solved with the PnP algorithm proposed by \cite{lepetit2009ep}. The reprojection error ($\varepsilon_{\mathcal{Y}_j}$) is then calculated with:
	\begin{align}
		\label{eqn:RSE_Rreproj}
		\varepsilon_{\mathcal{Y}_j} = 
		\sum_{i = 1}^{p} {\| \mathbf{K} \mathbf{T}(\mathbf{x}_{j}) \mathcal{F}_{i}
		- \mathcal{C}_{i} \|_{\scriptstyle 2}}^{2}.
	\end{align}

	\noindent $\mathbf{K}$ is, again, the intrinsic model of the camera. $\mathbf{T}(\cdot)$ is a mapping function that maps $\mathbf{x}$ to $\mathbb{SE}(3)$. The order of $\mathcal{C}_i$ depends on $\mathcal{Y}_j$. $\varepsilon_{\mathcal{Y}_j}$ with the minimum value implies the correct correspondence pair $\mathcal{P}$, which is the final return of the initialization module. Algorithm \ref{algo:RSE_INIT} summarizes the initialization process.

	\begin{algorithm}
		\DontPrintSemicolon
		\caption{Initialization}
		\label{algo:RSE_INIT}
		\SetKwFunction{sizeof}{sizeof}
		\SetKwFunction{null}{null}
		\SetKwFunction{false}{false}
		\SetKwFunction{true}{true}
		\SetKwFunction{getPCD}{getPCD}
		\SetKwFunction{getMAD}{getMAD}
		\SetKwFunction{hsvimg}{hsvimg}

		\SetKwFunction{isNeighborHue}{isNeighborHue}
		\SetKwFunction{GenPermute}{GenPermute}
		\SetKwFunction{EPnP}{EPnP}
		\SetKwFunction{getReprojError}{getReprojError}
		\SetKwFunction{HSV}{HSV}

		\SetKwFunction{KmeansCluster}{KmeansCluster}


		\SetKw{Continue}{continue}

		\SetKwFunction{FMain}{Init}
		\SetKwProg{Fn}{Function}{:}{end}
		\Fn{\FMain{$\mathcal{F}$, $\mathcal{C}$, \hsvimg}}
		{
			\If{\sizeof{F} $\neq$ \sizeof{C}}
			{$\mathcal{P}$ = \null \;
			\KwRet $\langle \false, \mathcal{P} \rangle$ \;} 
			\
			
			$\mathcal{PCD} \leftarrow$ \getPCD{$\mathcal{C}$} \;
			$\Delta_{\mathbf{MAD}} \leftarrow$ 
				\getMAD{$\mathcal{F}$} - \getMAD{$\mathcal{MAD}$} \;
			\If{$\Delta_{\mathbf{MAD}} > \lambda_{\mathbf{MAD}}$}
				{
					$\mathcal{P}$ = \null \;
					\KwRet $\langle \false, \mathcal{P} \rangle$ \;
				} 
			\

			$\mathcal{Y}\leftarrow$ \GenPermute{$\mathcal{C}$}\;
			$\varepsilon' \leftarrow  \inf$\;
			
			\For{$\mathcal{Y}_{j}$ in $\mathcal{Y}$ }
			{
				$\mathbf{hsv}_{c}\leftarrow$ \HSV{$\mathcal{Y}_{j}$, \hsvimg}\;
				\eIf{\isNeighborHue
					{
						$\mathbf{hsv}_{f}$,
						$\mathbf{hsv}_{c}$
					}
				}
				{						
					$\mathbf{x}_j \leftarrow$ 
					\EPnP{$\mathcal{F}$, $\mathcal{Y}_{j}$} \;

					$\varepsilon_{\mathcal{Y}_j} \leftarrow$  
					\getReprojError{
						$\mathbf{x}_m$, 
						$\mathcal{F}$,
						$\mathcal{Y}_{j}$
					}\;

					\If{$\varepsilon_{\mathcal{Y}_j} < \varepsilon'$}
					{
						$\mathcal{P} \leftarrow 
						\langle \mathcal{F}, \mathcal{Y}_{j} \rangle$ \;
						$\varepsilon' \leftarrow \varepsilon_{\mathcal{Y}_j} $						
					}
				}
				{
					\Continue \;
				}					
			}
			
			\KwRet $\langle \true, \mathcal{P} \rangle$ \;
		}
	\end{algorithm}
	
	\subsection{Correspondence Tracking}
	\label{sec:RSE_CT}
	After initialization, to preserve stable PnP measurement, the successive correspondences at $t>0$ should be retrieved correctly. At each time step $k$ ($k\neq$0), based on the estimated pose of $\mathbf{x}_{k-1}$ at time $k-1$, we reproject the points in $\mathcal{F}$ onto the image frame, which are denoted as $\mathcal{F}_{2\text{D}}$. Based on the maximum/minimum 2D values of $\mathcal{F}_{2\mathbf{D}}$, a bounding box $\mathbf{ROI}$ is generated. With the image filtered by $\mathbf{ROI}$ mask and depth information $\mathbf{s}_k$, we repeat eqn. \eqref{eqn:RSE_thres} to generate $\mathcal{C}_k$.

	As $\mathcal{F}_{2\mathbf{D}}$ is directly generated from $\mathcal{F}$, correspondence can be directly retrieved by conducting nearest neighbor with $\mathcal{F}_{2\mathbf{D}}$ and $\mathcal{C}_k$. Nevertheless, in the proposed system, as aggressive maneuvers can occur during landing, the displacement of LED blobs on a 2D images could be large. Therefore, performing the nearest neighbor search directly with $\mathcal{F}_{2\text{D}}$ derived from pose from $k-1$ could lead to wrong results. In \cite{faessler2014monocular}, a linear velocity model is used to predict the coordinates of new 2D blobs. However, such a method might not be stable during large acceleration or when the cluster of blobs is rather small on the image frame due to the relatively large distance between the camera and the quadrotor. Hence, a trivial but efficient modification is applied, in which $\mathcal{F}_{2\text{D}}$ is shifted by the center difference of $\mathcal{F}_{2\text{D}}$ and $\mathcal{C}_k$. We denote the shifted blob coordinates as $\mathcal{F}_{2\text{D}}^{'}$. The nearest neighbor search is then conducted based on $\mathcal{F}_{2\text{D}}^{'}$ and $\mathcal{C}_k$; the $\mathcal{P}$, along with the detected candidates $\mathcal{C}_k$, is then the final measurement output of correspondence tracking.

	\subsection{Recursive Filtering - Predict}
	\label{sec:RSE_RF_predict}
	A recursive filtering method is proposed to track the estimation of the relative states, where an Iterated Extended Kalman filter (IEKF) is adopted. As the state is expressed on manifold, as shown in eqn. \eqref{eqn:states}, we involve Lie algebra to perform exponential and logarithm mapping when deriving derivatives and integrals.

	The non-linear prediction model harnessed here can be first written as:
	\begin{align}
		\label{eqn:RSE_predict1}
		\dot{\mathbf{p}} &= \mathbf{v} \nonumber \\
		\dot{\mathbf{R}} &= \mathbf{R}{\boldsymbol{\xi}_{1}}_{\times} \nonumber \\
		\dot{\mathbf{v}} &= \mathbf{R}(-\mathbf{g} + \boldsymbol{\xi}_{2}) + \mathbf{g}.
	\end{align}
	
	\noindent 
	From the above equations, $\mathbf{g}$ is $[0,0,-9.81]^T \in \mathbb{R}^3$. As neither IMU data nor control input are used during estimation, the model is simplified and premised with assumptions. In which, inspired by \cite{fu2017robust}, we omit the angular acceleration term, and model the angular velocity as Gaussian noises, $\boldsymbol{\xi}_{1} \in \mathbb{R}^3$. Additionally, the acceleration induced by control input is approximated as $-\mathbf{g} + \boldsymbol{\xi}_{2}$, which is assumed to be close to hover thrust. $\boldsymbol{\xi}_{2}$ is set to be $[0,0,\xi_z] \in \mathbb{R}^3$ with $\xi_z$ also being Gaussian noise. Also note that the term $-\mathbf{g} + \boldsymbol{\xi}_{2}$ is defined in the body frame. Therefore, the prediction stage is given as the following:
	\newcommand{\at}[2][]{#1|_{#2}}
	\begin{align}
		\label{eqn:RSE_predict4}
		{\mathbf{x}}_k^{-} &= \mathbf{x}_{k-1} \boxplus \delta t \delta \mathbf{x}, \nonumber\\
		where \;\; \delta \mathbf{x} &= f(\mathbf{x}_{k-1}, \mathbf{w}).\\
		\label{eqn:RSE_predict7}
		\mathbf{P}^{-}_k &=  \mathbf{F_k P_{k-1}F_k}^T + \mathbf{Q}, \nonumber\\
		where \;\; \mathbf{F}_k &= \frac{\partial f}{\partial \mathbf{x}}\Bigr|_{\mathbf{x} = \mathbf{x}_{k-1}}.
	\end{align}

	\noindent
	The nonlinear function defined in eqn. \eqref{eqn:RSE_predict4} is based on eqn. \eqref{eqn:RSE_predict1}, and eqn. \eqref{eqn:RSE_predict4} is constructed assuming a constant-velocity motion model. Note that $\boxplus$ operation is defined as
	\begin{align}
		\mathbf{A} \boxplus \Delta = \mathbf{A} \mathbf{Exp}(\Delta), 
		\mathbf{A} \in \mathcal{M}, \Delta \in \mathbb{R}^9,
	\end{align}
 
	\noindent where $\mathbf{Exp}(\cdot)$ is the exponential mapping that maps the Lie algebra of $\mathcal{M}$, i.e., $\mathfrak{m} = \mathfrak{r} \oplus \mathfrak{so}(3) \oplus \mathfrak{r}$, to $\mathcal{M}$ (defined in eqn. \eqref{eqn:states}).
	Hence, $\Delta$ is defined within the tangent space ($\mathbb{R}^9$) of the Lie group. Furthermore, in eqn. \eqref{eqn:RSE_predict7}, the covariance matrix $\mathbf{P}^{-}_k$ is defined as the local perturbations around $\mathbf{x}_k \in \mathcal{M}$, whereas the $\mathbf{F}_k$ is the derived Jacobian matrix on the Lie group for covariance propagation.

	\subsection{Recursive Filtering - Update}
	\label{sec:RSE_RF_update}
	The main difference between EKF and IEKF is how the filter updates the state based on priori and measurement, whereas the latter iteratively performs linearization and updates accordingly. Here, we linearize the pin-hole camera model. The main motivation for using IEKF is to alleviate the linearization errors induced in conventional EKF, determining the state as a maximum a posteriori (MAP) estimate. Particularly, we follow the method of \cite{bell1993iterated}, in which the posteriori state is returned by performing Gauss-Newton optimization. 

	During the update stage, in addition to the measurement for pose inference (sec. \ref{sec:RSE_CT}), we also retrieve velocity measurement based on a constant-velocity model by setting $\mathbf{v_z}_k$ as $(\mathbf{p}_{k-1} - \mathbf{p}_{k-2})/\delta t_k$. Nevertheless, as it is derived directly from numerical differentiation, the returned signal is intermingled with high-frequency noise; therefore, we preprocess it with a low-pass IIR filter:
	\begin{align}
		\mathbf{v_z}_k \leftarrow \alpha * \mathbf{v_z}_k + (1 - \alpha) * \mathbf{v_z}_{k-1},
	\end{align}
 
	\noindent where $\alpha \in [0,1]$. Subsequently, the update step could be then summarized as below in eqn. \eqref{eqn:IEKF_GN}. Note that time step $k$ is abbreviated here for readability.
	\begin{align}
		\label{eqn:IEKF_GN}
		\mathbf{x}_k &= \arg \min_{\mathbf{x}}\{
			r_{\mathcal{D}}(\mathbf{x},{\mathbf{x}}^{-}, \mathbf{P}^{-})
			+ 
			r_{\mathcal{V}}(\mathbf{x}, \mathcal{C}, \mathbf{v_z}, \mathbf{R})
		 \} \\
		 r_{\mathcal{D}}(\cdot)
		 &= \| \mathbf{x} \boxminus \mathbf{x}^- \|^{2}_{\mathbf{P}^{-}} \\
		 r_{\mathcal{V}}(\cdot)
		 &= \| \mathbf{V}(\mathbf{x}) - \mathbf{v}\|^{2}_{\mathbf{R}_{v}} +
		 \sum_{i=0}^{q} \| \mathbf{K} \mathbf{T}(\mathbf{x})\mathcal{F}_i - \mathcal{C}_i  \|^{2}_{\mathbf{R}_{p}}.
	\end{align}

	The non-linear least squared function consists of two residual functions, $r_{\mathcal{D}}$ and $r_{\mathcal{V}}$, which are respectively dynamic factor and visual factor for the single-time MAP problem. Moreover, $\mathbf{P}^{-}$ is the priori covariance matrix, and $\mathbf{R}$ is the measurement noise matrix. As for $\mathbf{V}(\cdot)$, it is the mapping function that maps $\mathbf{x}$ to $\mathbb{R}^3$, which is the velocity vector space. As mentioned, optimal $\mathbf{x}_k$ is solved via Gauss-Newton optimization. Also note that $\boxminus$ operation is defined as:
	\begin{align}
		\mathbf{A} \boxminus \mathbf{B} = 
		\mathbf{Log}(\mathbf{B}^{-1} \circ \mathbf{A}) = \Delta, 
		\mathbf{A}, \mathbf{B} \in \mathcal{M}, \Delta \in \mathbb{R}^9,
	\end{align}

	\noindent where the $\mathbf{Log}(\cdot)$ is the logarithm mapping, inverse mapping of $\mathbf{Exp}(\cdot)$. As for $\circ$, it is the group composition operation within the Lie group.
	
	Afterwards, we also propagate the covariance information after all iterations by calculating the Kalman gain based on the final $\mathbf{x}_k$:
	\begin{align}
		\mathbf{H}_k &= \frac{\partial h}{\partial \mathbf{x}}\Bigr|_{\mathbf{x} = {\mathbf{x}}_k} \nonumber,\\
		\mathbf{S}_k &= \mathbf{H}_k \mathbf{P}^{-}_k \mathbf{H}^{T}_k + \mathbf{R} \nonumber,\\
		\mathbf{K}_k &= \mathbf{P}^{-}_k \mathbf{H}^T_k \mathbf{S}^{-1}_k \nonumber,\\
		\mathbf{P}_k &= (\mathbf{I} - \mathbf{K}_k \mathbf{H}_k)\mathbf{P}_k^{-}
			(\mathbf{I} - \mathbf{K}_k \mathbf{H}_k)^T + 
			\mathbf{K}_k \mathbf{R} \mathbf{K}_k^T.
	\end{align}
 
	\noindent 
	Note that for final covariance propagation, we adopt the Joseph formulation \cite{brown1997introduction} for numerical stability so that matrix symmetry can be ensured. The final retrieved state $\mathbf{x}_k$ is then forwarded as estimation feedback for motion planning.

\section{Landing Motion Planning}
\label{sec:motion}
	This section elucidates the motion planning module. To begin with, we program the motion primitive in a relative fashion, meaning that the generated trajectories are coupled with the relative states and relative constraints of the quadrotor with respect to the local frame ($\mathcal{N}$) of the dynamic platform. Furthermore, to ensure the quadrotor stays within the FoV at all times, the trajectory generator takes visibility-safety into consideration and imposes kinematic constraints accordingly during the optimization stage. Additionally, the Bézier basis is harnessed to guarantee the aforementioned constraints, while time allocation for the trajectory is further discussed to improve the optimality of the result. 

	\subsection{Motion Planning in Relative Local Frame}
	\label{sec:MP_RLF}
	The motion planner resonates extensively with classic trajectory optimization methods \cite{mellinger2011minimum}, where the property of differential flatness is exploited. A system's dynamic is considered differential flat when the state and the inputs could be derived from a set of \emph{flat outputs} and their derivatives. Like many trajectory generation literatures, we utilized the Cartesian translation vector and yaw, i.e., $\sigma_\mathcal{N} = [\mathbf{x},\mathbf{y},\mathbf{z}, \psi ]^{T}$, as flat outputs.
	Therefore, equation eqn. \eqref{eqn:SO_traj} is reformulated with flat outputs and with subscript $\mathcal{N}$:
	\begin{align}
		\label{eqn:MP_traj}
		\boldsymbol{\sigma}_{\mathcal{N}}(t) = 
		(	x_\mathcal{N}(t), 
			y_\mathcal{N}(t), 
			z_\mathcal{N}(t), 
			\psi_\mathcal{N}(t)
		),  
		\forall t \in [T_0,T].
	\end{align}

	As the trajectory is generated within the local frame, to ensure dynamic feasibility, $\boldsymbol{\sigma}_\mathcal{N}(t)$ in eqn. \eqref{eqn:MP_traj} should be confined accordingly where the constraints are expressed locally in $\mathcal{N}$. Therefore, dynamic constraints $\boldsymbol{\beta}_{D} \in \mathbb{R}^6$, which comprises velocity and acceleration at time $t$ during optimization, could be retrieved by,
	\begin{align}
		\label{eqn:MP_dynconstr}
		^{\mathcal{N}}\boldsymbol{\beta}_{D}(t) &= 
		[{\mathbf{R}^{\mathcal{N}}_{\mathcal{I}}}(t) \oplus {\mathbf{R}^{\mathcal{N}}_{\mathcal{I}}}(t)]
		[^{\mathcal{I}}\boldsymbol{\beta}_{D} - ^{\mathcal{I}}\boldsymbol{\beta}_{N}(t)], \nonumber \\
		where \;\;
		\mathbf{R}^{\mathcal{N}}_{\mathcal{I}} &\in \mathbb{SO}(3),\nonumber \\
		^{\circledcirc}\boldsymbol{\beta}_{\circledcirc} &\in \mathbb{R}^{6}.
	\end{align}

	\noindent $\mathbf{R}^{\mathcal{N}}_{\mathcal{I}}$ rotates translational vectors defined in inertial frame $\mathcal{I}$ to non-inertial target frame $\mathcal{N}$; in this case, we apply the direct sum of two identical rotation matrices to rotate the dynamic vector $^{\circledcirc}\boldsymbol{\beta}_{\circledcirc}$. $^{\mathcal{I}}\boldsymbol{\beta}_{D}$ is the dynamic constraints of the quadrotor defined in $\mathcal{I}$, whereas $^{\mathcal{I}}\boldsymbol{\beta}_{N}$ is the velocity and acceleration of the target frame, i.e., the moving platform defined in $\mathcal{I}$. By setting the premise of motion planning in the local frame, we further detail the trajectory generation optimization problem. 

	\subsection{Visibility-Safety Corridor Generation}
	\label{sec:MP_VSFC}
\begin{figure}[hbt!]		
		\centering
		\includegraphics[width=0.5\textwidth]{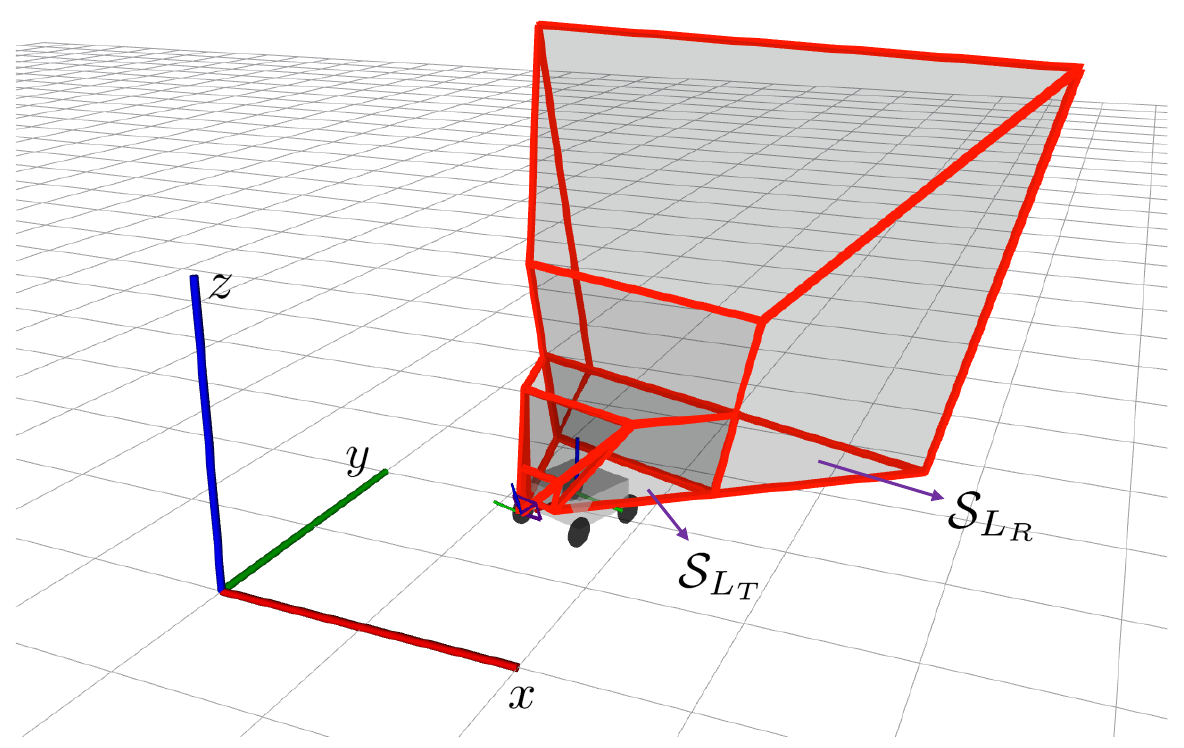}
		\caption{A graphic presentation of the proposed VSFC $\mathcal{S}_L$, which consists two subsets, \textit{rear} space $\mathcal{S}_{L_R}$ and \textit{touchdown} space $\mathcal{S}_{L_T}$. Visualization package developed by \cite{liu2017planning} is used here.}
		\label{fig:VSFC}
		\vspace{-0.0cm}	 
	\end{figure}

	In the proposed system, to ensure visibility, it is crucial to constrain the quadrotor within the limited FoV of the camera at all times during the landing stage. To achieve this, we utilize convex polyhedra to construct the safe flight corridor denoted as the visual safe flight corridor (VSFC). Note that VSFC is also expressed in local frame $\mathcal{N}$.
	\begin{enumerate}
		\item Upon initialization of the recursive filter tracker (sec. \hyperref[sec:RSE]{3}), a landing cubic free space ($\mathcal S$) is generated based on the initial state ($\mathbf{x}_{0}$) and terminate state ($\mathbf{x}_{T}$), where $\mathbf{x}_{T_0}, \mathbf{x}_{T} \in \mathcal S$. $\mathcal{S}$ also excludes occupancies $\mathcal{O}$, which constitutes the moving landing platform and the surroundings, and hence, $\mathcal{S} \leftarrow  \mathcal{S} \setminus \mathcal{O}$.
		
		\item VSFC generator first segment $\mathcal{S}$ into two: \emph{rear} space ($\mathcal{S}_{R}$) and \emph{touchdown} space ($\mathcal{S}_{T}$). Specifically, we confine $\mathcal{S}_{T}$ with a touch down height ($h_{T}$) and exclude the space above $h_{T}$. This is to reduce ground-effect at touchdown, where $\mathcal{S}_{T}$ is deflated to make the trajectory have less vertical maneuver. $\mathcal{S}_{R}$ and $\mathcal{S}_{T}$ are then enlarged with dilate coefficient $\epsilon_{d}$ to guarantee the continuity of the entire landing space so that $\mathcal{S}_{R} \cap \mathcal{S}_{T} \neq \emptyset$. 

		\item We then utilize the FoV information of the camera, i.e., $\theta_{w} \times \theta_{h}$, along with sensor translation and rotation in local frame to construct tangent hyperplanes; each hyperplane induces a halfspace, the set of halfspaces is then expressed as:
		\begin{align}
			\label{eqn:MP_HS}
			\mathcal{S}_C  = \{\mathbf{s} | \mathbf{a}^{T}_{i} \boldsymbol{\sigma} \leq \mathbf{b}_{i}, 
		 	\forall i = 0, 1, 2... ,l\}.
		\end{align}

		\item Therefore, the final VSFC could be expressed as convex polyhedra,
		\begin{align}
			\label{eqn:MP_KINE}
			\mathcal{S}_L = \{
				\mathcal{S}_{L_i} | 
				\mathcal{S}_{L_i} = \mathcal{S}_{i} \cap \mathcal{S}_C,
				i = R, T
			\}. 
		\end{align}

		From eqn. \eqref{eqn:MP_KINE}, convexity is preserved as each polyhedron is the intersection of convex halfspaces. $i$ denotes the two consecutive landing polyhedra, namely, \emph{rear} and \emph{touchdown}. Lastly, to guarantee safety, the robot is modeled as a sphere with radius $r_{q}$; we then shrink $\mathcal{S}_L $ with $\epsilon_{s}$, $\mathcal{S}_L  \leftarrow \epsilon_{s} \mathcal{S}_L $. Figure \ref{fig:VSFC} exemplifies the defined VSFC.
	\end{enumerate}

	\subsection{Trajectory Generation based on Bézier Basis}
	\label{sec:MP_TG}
	A smooth trajectory is then planned as piecewise polynomials, in which flat outputs defined in eqn. \eqref{eqn:MP_traj} are utilized and parameterized with $t$. Concurrently, the generated trajectory should be subject to constraints defined in sections \ref{sec:MP_RLF} and \ref{sec:MP_VSFC}.
	
	A common polynomial expressed with monomial bases is usually structured as
	\begin{align}
		\label{eqn:MP_polymono}
		\mathbf{p}(t) = 
		\sum_{i = 0}^{n} \mathbf{p}_{i}t^{i},
	\end{align}

	\noindent which is an $n$-th order polynomial. In the proposed system, we modify the conventional polynomial to Bézier curve, a polynomial represented by Bernstein polynomials bases. The main motivation for such reformulation is to ensure the motion primitives stay within the dynamic and kinematic feasible region; this method has been validated by 
	\cite{honig2018trajectory}.
	First, the original polynomial is recast as, 
	\begin{align}
		\label{eqn:MP_Bernstein}
		&\mathbf{p}_{\mathbb{B}}(t) = 
		\sum_{i = 0}^{n} \mathbf{c}^{i} b^{i}_{n}(t), \nonumber\\
		where \; \; &b^{i}_{n}(t) = {n \choose i}t^{i}(1-t)^{n-i},
	\end{align}

	\noindent in which the $n$-degree curve is defined by Bernstein basis $b^{i}_{n}(t)$ along with a series of control points $\mathbf{c}^{i}$. Through utilizing control points, Bézier curve yields two useful properties to allow trajectories to comply with some hard constraints \cite{riesenfeld1973applications}:

	\begin{enumerate}
		\item \emph{Convex Hull} Property. Based on the control points $\mathbf{c}^{i}$, the generated trajectory $\mathbf{p}_{\mathbb{B}}(t)$ is confined within the convex hull constructed by $\mathbf{c}^{i}$'s. Therefore, by restricting the feasible region of control points based on VSFC eqn. \eqref{eqn:MP_KINE}, the kinematic feasibility could be ensured.
		
		\item \emph{Hodograph} Property. With the hodograph property, the derivatives of the Bézier curve is also a Bézier curve, which could be expressed in terms of another set of $\mathbf{c}^{i'}$, which can formulate another Bézier curve defined in eqn. \eqref{eqn:MP_Bernstein}. Therefore, by utilizing this characteristic, the dynamic feasibility is guaranteed, where they are confined by eqn. \eqref{eqn:MP_dynconstr}.
	\end{enumerate}

	Meanwhile, to generate an optimal trajectory, we minimize the integral of the squared $3^{rd}$-derivative (i.e., the jerk) of the trajectory with respect to the weights, i.e., control points $\mathbf{c}^{i}$, of the trajectory. The objective cost function is shown below first:
	\begin{align}
		\label{eqn:MP_COST}
		\mathbf{J} = \int_{T_{0}}^{T}\left\| 
			\frac{d^{3}\mathbf{p}_{\mathbb{B}}(t)}{dt^{3}} 
			\right\|^{2}\;dt.
	\end{align}

	\noindent We then re-write the objective in terms of optimization variables($\mathbf{c}$), mapping ($\mathbf{M}$), and differential matrices ($\mathbf{Q}$).
	\begin{align}
		\label{eqn:MP_COSTMAT}
    \mathbf{J} = \int_{T_{0}}^{T}\left\| 
			\frac{d^{3}\mathbf{p}_{\mathbb{B}}(t)}{dt^{3}} 
			\right\|^{2}\;dt 
      = \mathbf{c}^T \mathbf{M}^T
      \mathbf{Q} \mathbf{M} \mathbf{c}
	\end{align}

	\noindent As mentioned, we also want the trajectory to comply with feasibilities, and hence, the optimization problem is subject to affine constraints, as stated in eqn. \eqref{eqn:MP_dynconstr} and eqn. \eqref{eqn:MP_KINE}, along with equality constraints induced by starting and terminating state, and continuity between each polynomial. The final optimization problem is hence formulated as follows: 
	\begin{align}
		\label{eqn:MP_OPT}
		\nonumber \underset{\mathbf{c} \in \mathbb{R}^{3} }{\text{minimize}} &
		\;\;\mathbf{c}^T \mathbf{M}^T \mathbf{Q} \mathbf{M} \mathbf{c} \\
		\nonumber \text{subject to} & \;\;\\
		\nonumber \Lambda_{eq} \mathbf{c} &=  \boldsymbol{\beta}_{eq}\\
		\nonumber \Lambda_{D} \mathbf{c} &\preceq  \boldsymbol{\beta}_{D}\\
				  \Lambda_{S} \mathbf{c} &\preceq  \boldsymbol{\beta}_{S},
	\end{align}
	
	\noindent in which $\Lambda_{eq}$ denotes the affine relation matrix between control points $\mathbf{c}$ and equality constraints, while $\Lambda_{D}$ and $\Lambda_{S}$ map $\mathbf{c}$ to inequality constraints of eqn. \eqref{eqn:MP_dynconstr} and eqn. \eqref{eqn:MP_KINE}. From eqn. \eqref{eqn:MP_OPT}, one could easily identify it as a convex problem, as it has a convex objective function and convex constraints. Furthermore, it is also a quadratic programming (QP) problem, which can be easily solved by off-the-shelf QP solvers.
	
	\subsection{On Time Allocation}
	\label{sec:MP_TA}
	From eqn. \eqref{eqn:MP_OPT}, it is seen that the derivative matrix $\mathbf{Q}$ is temporally dependent, meaning that $\mathbf{Q}$ changes with different time inputs. Therefore, this implies that with different time allocations, the objective of the optimization problem would differ, leading to suboptimal results. To achieve better results, we sample a sequence of time allocations and evaluate each cost accordingly within the set. Below is the proposed strategy step by step:

	\emph{(1) Sample on start}.
		Once the guidance planner is initialized, a time allocation $\mathcal{T}$ set is constructed, in which each element is sampled within the region of minimum and maximum time calculated based on the velocity in dynamic constraints, namely the best and worst case scenarios.

	\emph{(2) Constraints tightening}.
		From eqn. \eqref{eqn:MP_OPT} and eqn. \eqref{eqn:MP_dynconstr}, it could be readily identified that solving the trajectory online is a perturbed problem \cite[p. 249]{boyd2004convex}. Therefore, it is natural to tighten the constraints when optimizing throughout the set so a feasible solution exists during online optimization. Consequently, for eqn. \eqref{eqn:MP_dynconstr} during sampling stage, minimum $\boldsymbol{\beta}_{\mathcal{N}}$ and maximum $\boldsymbol{\beta}_{\mathcal{P}}$ are selected to retrieve the most conservative dynamic constraints.
		
	\emph{(3) Optimization throughout the set}.
		\label{subsec:MP_OPT} Each sample time is then visited, and its respective optimization is set up and solved. The problem with the lowest returned cost will yield the best trajectory segment times. The planner solver will then take the optimal time allocation to formulate the objective function.
		
	\emph{(4) Online solving}.
		As mentioned, the returned time allocation will serve as the optimal segment times. At the landing stage, the planner will use this to construct the $\mathbf{Q}$ matrix accordingly and solve the objective as shown in eqn. \eqref{eqn:MP_OPT}.

	\subsection{Feedback Control}
	\label{sec:MP_control}
	Lastly, a trajectory parameterized with time $t$ is solved during the landing stage. Based on the polynomial, a series of setpoints could be calculated; with each setpoint at time step $t_k$ and state $\mathbf{x}_k$, a Propotional-Integral-Derivative (PID) controller will then infer the corresponding outer-loop control command. The onboard attitude controller will then track the control signal, which is transmitted to the quadrotor from the ground. 

\section{Experimental Results and Discussion}
\label{sec:results}

\subsection{Experiment Setup}
\begin{figure}[b]		
		\centering
		\includegraphics[width=0.5\textwidth]{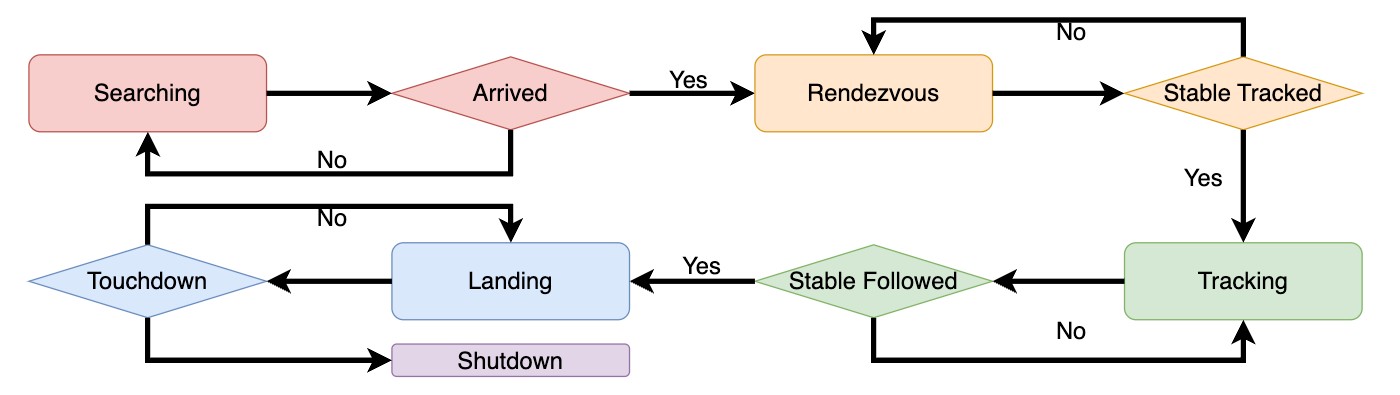}
		\caption{The predefined finite state machine.}
		\label{fig:fsm}
		\vspace{-0.0cm}	 
	\end{figure}

\begin{figure}[b]		
		\centering
		\includegraphics[width=0.45\textwidth]{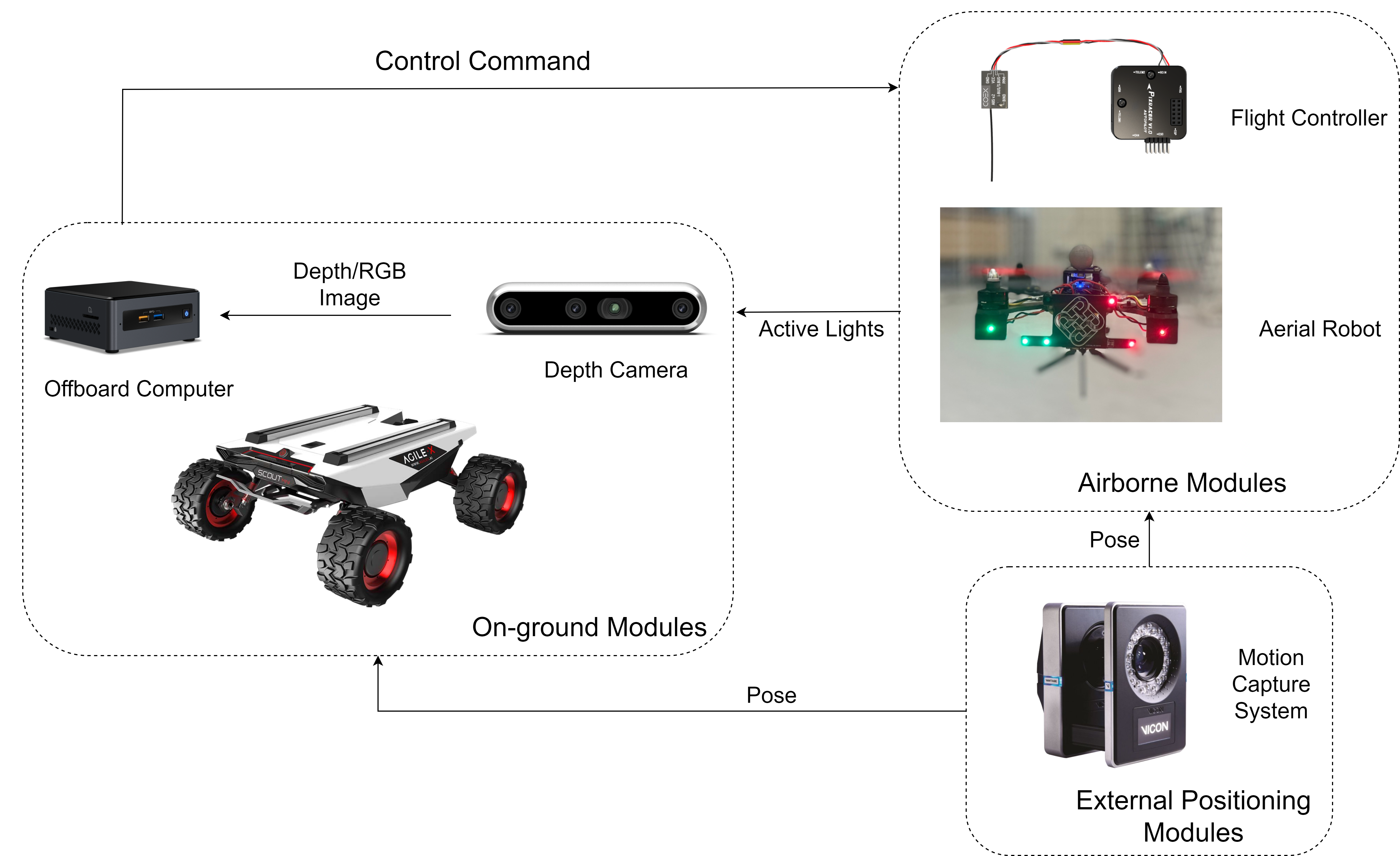}
		\caption{Experimental setup for the proposed system.}
		\label{fig:setup}
		\vspace{-0.0cm}	 
	\end{figure}

	Experiments presented in this section are done in a controlled environment installed with motion capture system (VICON); the overall landing mission is guided with a pre-defined state machine, as shown in Figure \ref{fig:fsm}. Figure \ref{fig:setup}, on the other hand, depicts the experimental setup.
	In the proposed work, we validate the design with a minimalistic setup. Specifically, for the quadrotor, we tailor-cut a carbon fiber quadrotor frame with a dimension of 18 cm \texttimes \space 18 cm and installed a Pixracer controller hardware and a WIFI-equipped MCU module on it. The controller hardware, with PX4 \footnote{https://px4.io/} firmware embedded, provides inner-loop control, whereas the WIFI module is used to receive control signals from the ground. In addition to above, we install LED markers at the front of the quadrotor for the pose estimation module, which is mounted with a configuration that avoids symmetry and co-planarity so that degenerated measurement can be prevented.
	
	Meanwhile, for on-ground modules, AgileX Scout Mini \footnote{https://global.agilex.ai/products/scout-mini} is opted for the main platform, on which we place the landing pad (43 cm \texttimes \space 43 cm), depth camera (Intel RealSense D455), and a minicomputer (Intel NUC8i7BEH). We fixate the camera with an upward-facing $20^{\circ}$; it has a FoV of $87^{\circ} \times 58^{\circ}$.
	
    On the ground computer, several threads are run parallelly, where the following modules are executed: (1) pose estimation, (2) trajectory planner, (3) quadrotor outer-loop controller, and (4) ground vehicle path planner and controller. Furthermore, all packages are programmed in C++/Python within the framework of ROS. As for quadratic programming solving during motion generation, we use OSQP \cite{osqp}, which can provide efficient solutions within milliseconds and an easy interfacing module for integration.
 
	\subsection{Relative State Estimation}
	\label{sec:EXP_RSE}
	\begin{figure}[t]
		\centering
		\subfloat[][]
		{%
		\includegraphics[width=0.24\textwidth]{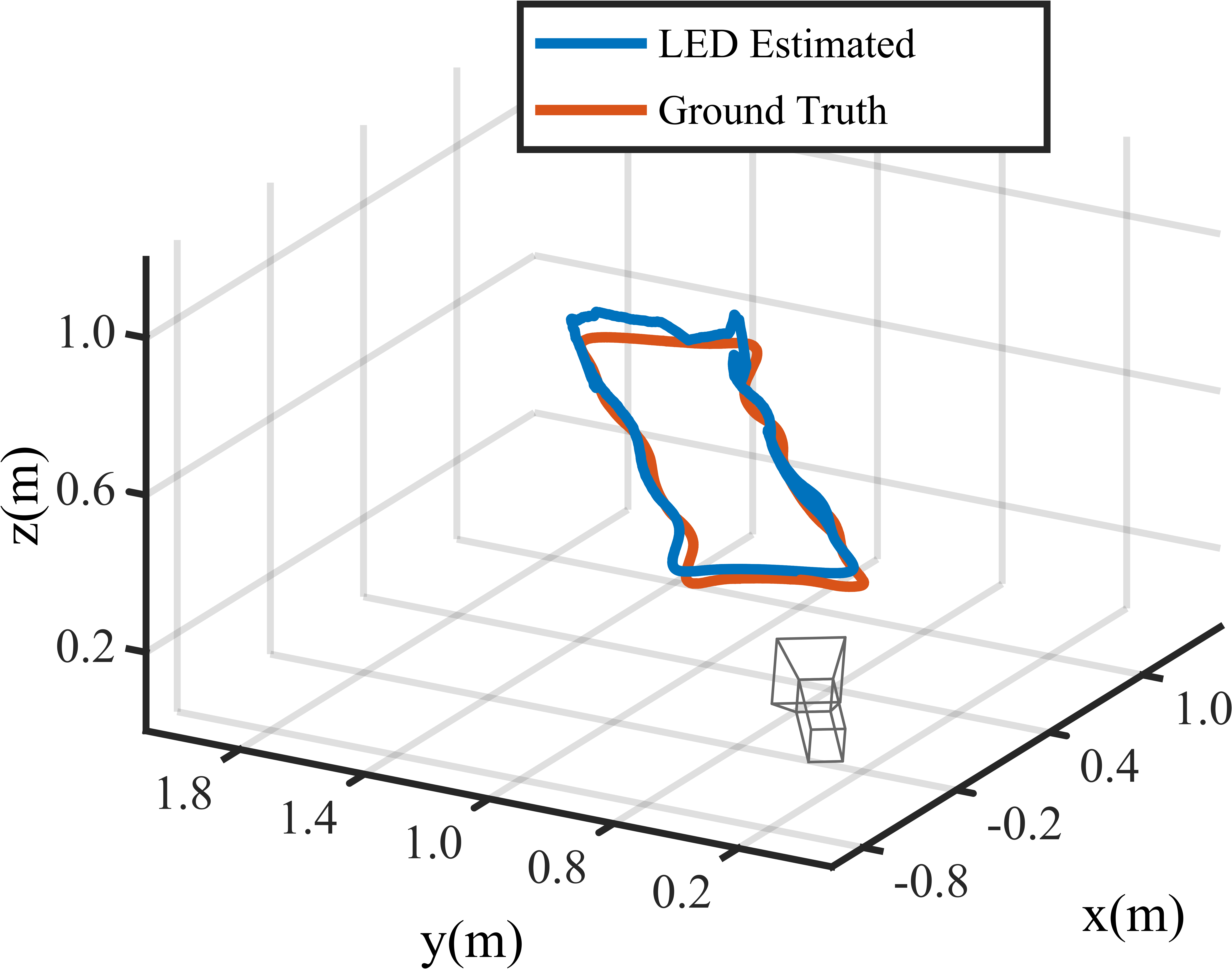}%
		\label{fig:rse_static_block}
		}%
		\hspace*{0.2cm}
		\subfloat[][]
		{%
		\includegraphics[width=0.24\textwidth]{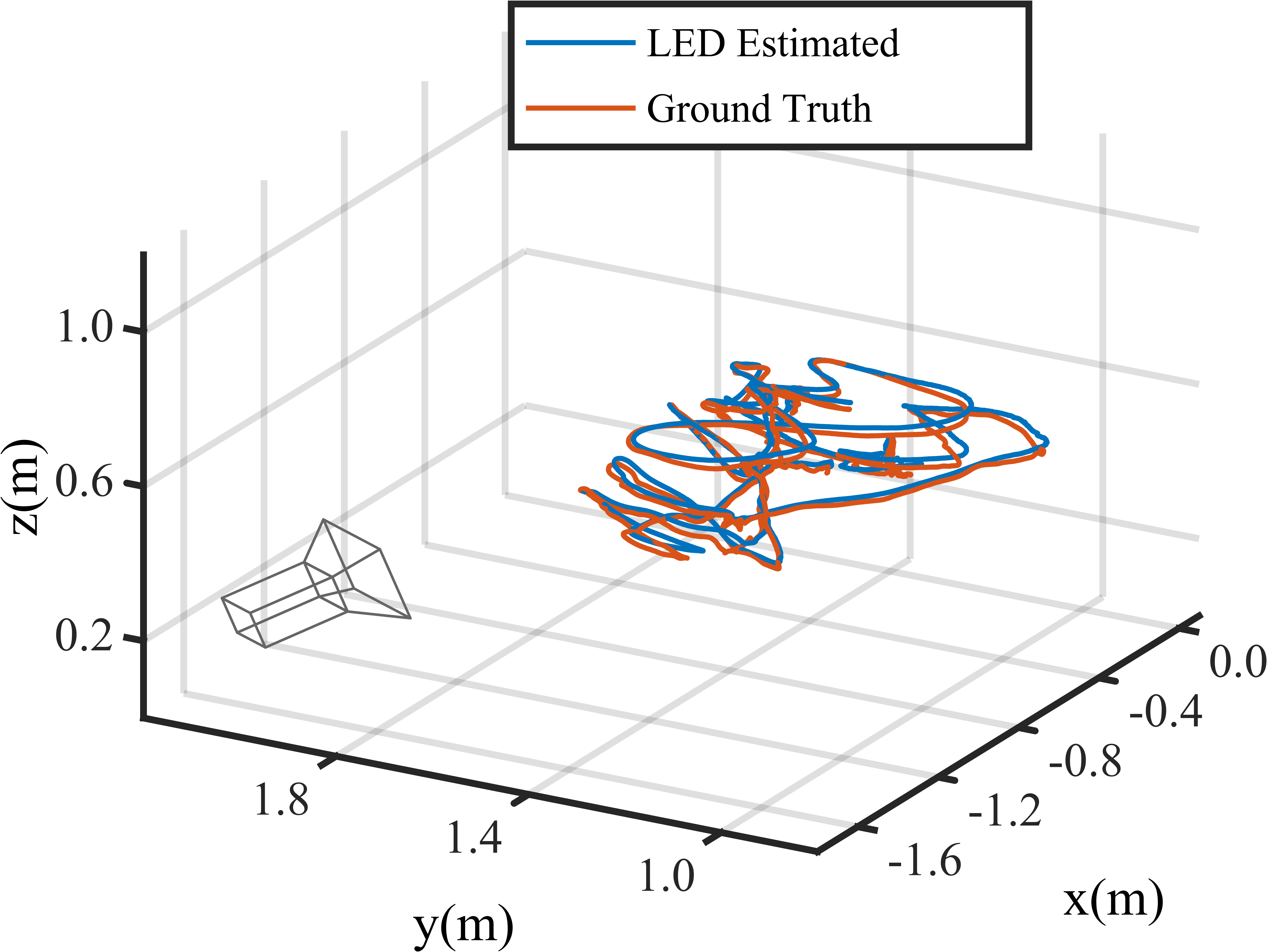}%
		\label{fig:rse_static_random}
		}
		\vspace*{0.1cm}
		\subfloat[][]
		{%
		\includegraphics[width=0.24\textwidth]{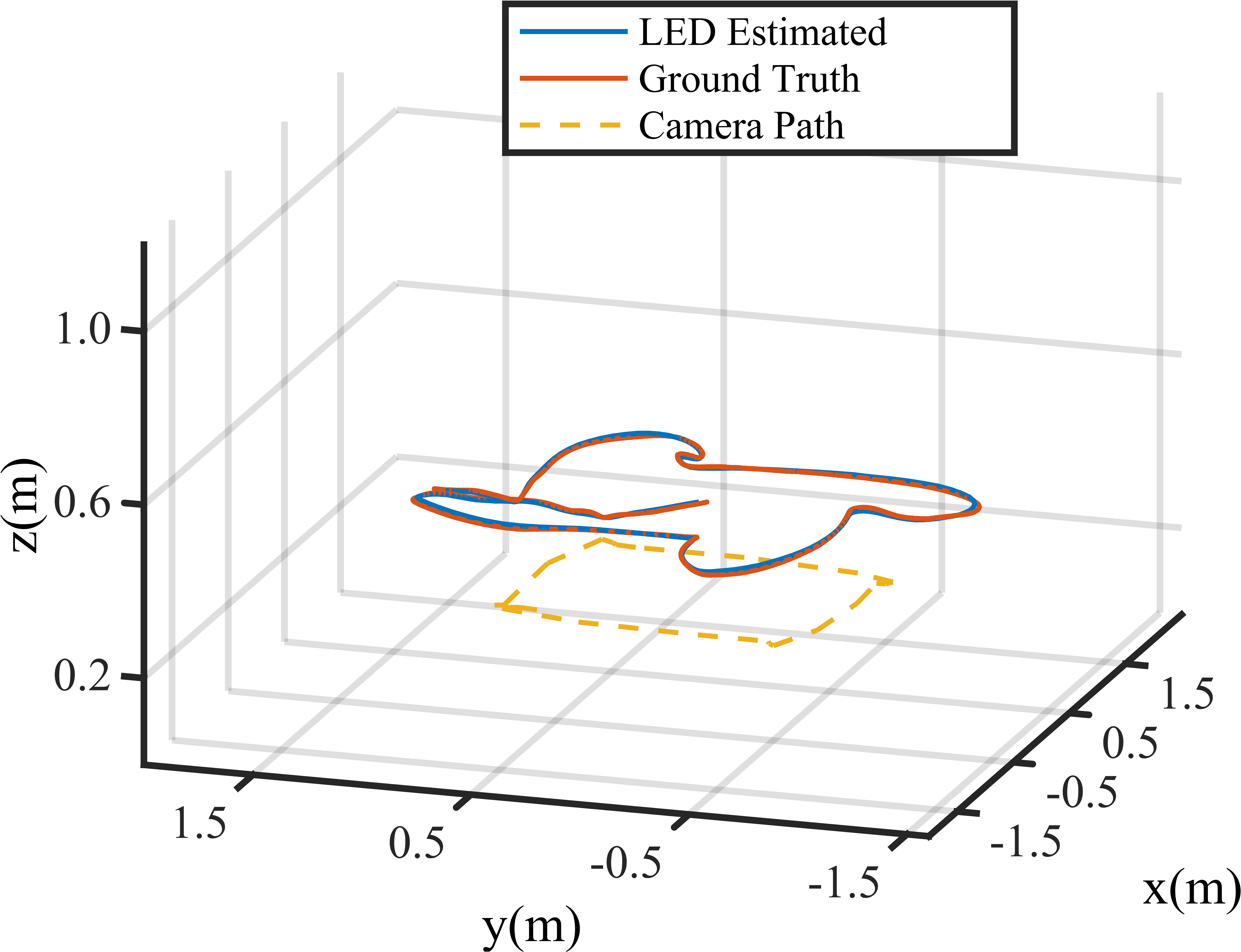}%
		\label{fig:rse_dynamic_block}
		}
		\hspace*{0.2cm}
		\subfloat[][]
		{%
		\includegraphics[width=0.24\textwidth]{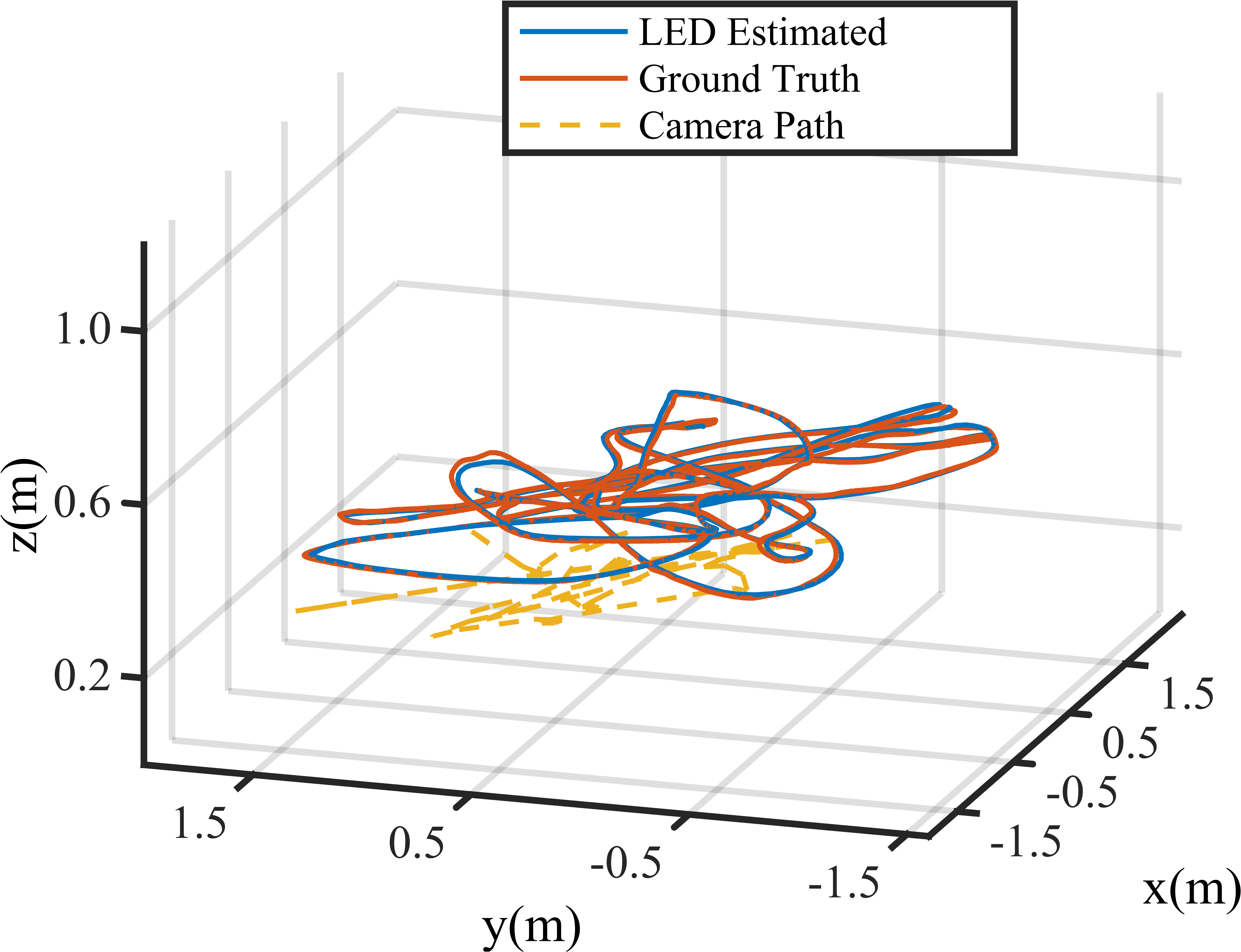}%
		\label{fig:rse_dynamic_random}
		}
		\caption{Relative state estimation.
		(a) Quadrotor moving in a block path in front of the camera.
		(b) Random maneuver in front of the camera.
		(c) Camera moving in block path, while quadrotor follows.
		(d) Camera moving in block path, while quadrotor flies in block path defined in the non-inertial frame $\mathcal{N}$.}
	\end{figure}
	The relative state estimation is assessed by the following: (1) direct evaluation with ground truth data when the camera is still, (2) direct evaluation with ground truth data when the camera is in motion, and (3) comparison with other relative state estimation methods. Note that the ground truth data ($\bar{\mathbf{x}}$) is collected from VICON, and we calculate the absolute trajectory error (ATE) as the evaluation metric:
	\begin{align}
		e_{ATE} = \frac{1}{N} \sum_{i=1}^{N}\sqrt{\frac{1}{K} \sum_{k=1}^{K}\|\mathbf{x}_{k,i} \boxminus \bar{\mathbf{x}}_{k,i} \|^2_2}.
	\end{align}
	\noindent We also calculate the tracking success rate, where we deem a tracking frame successful based on its reprojection error.

	\textit{(1) Camera at still:} 
	Figures \ref{fig:rse_static_block} and \ref{fig:rse_static_random} first showcase the recorded estimation and ground truth trajectories when the camera is fixed within the inertial frame $\mathcal{I}$. Specifically, in figure \ref{fig:rse_static_block}, we programmed a rectangular path within the non-inertial frame $\mathcal{N}$ for the UAV to follow. As for figure \ref{fig:rse_static_random}, we randomly controlled the UAV in front of the camera. Within the two experiments, the relative distance between the two varies from 0.4 $m$ to 3 $m$. From table \ref{tab:rse}, it can be seen that the designed system yields satisfactory results, where the position ATE is less than 0.03 $m$, whereas the orientation ATE is smaller than $4^{\circ}$ in the static case. Furthermore, the tracking success rate is 100 \%.

	\textit{(2) Camera at motion:} Datasets with the camera in motion were also collected, as shown in figures \ref{fig:rse_dynamic_block} and \ref{fig:rse_dynamic_random}. Specifically, in \ref{fig:rse_dynamic_block}, we programmed the UGV to move in a block path and let the quadrotor fix at a relative position defined in non-inertial frame $\mathcal{N}$; in \ref{fig:rse_dynamic_random}, UGV was controlled manually, and UAV was designed to move in a block path in $\mathcal{N}$. Table \ref{tab:rse} shows that the overall accuracy is within acceptable range for both translation and rotation, with position ATE less than 0.05 $m$ and the orientation ATE less than $5^{\circ}$. In addition, the tracking success rate is 99.78\%. Therefore, we deem the overall performance stable, whether the camera is static or dynamic. The module could, hence, provide a reliable state estimation feedback for our non-robocentric landing framework.

	\begin{table}[ht]
		\centering
		\caption{\underline{Results from Relative State Estimation}}
		\label{tab:rse}
		\centering
		\begin{tabular}{l|lll} 
							 & Static & Dynamic & Unit  \\ 
			\hhline{====}
			ATE\_translation & 0.021  & 0.049   & $m$     \\
			ATE\_rotation    & 3.501  & 4.727   & $^\circ$    \\
			Success Rate             & 100.00 & 99.78   & \%   
		\end{tabular}
	\end{table}

	\textit{(3) Comparison with others:} 
	We also compare the proposed IEKF method with other methods: iterative method \cite{faessler2014monocular} (denoted as KF-less) and ArUco. The ArUco tag is fixated on the drone, similar to the attached constellation. The comparison is carried out on several datasets with two scenarios: (a) quadrotor following UGV in a circle path and (b) quadrotor landing on a moving platform. Table \ref{tab:compare} reveals the performances of each method. From the table, it could be seen that IEKF returns better accuracy in both scenarios in terms of translation and rotation, where IEKF gives an average of 0.036 $m$ and 0.053 $m$ position error as well as $3.104^\circ$ and $3.007^\circ$ orientation error. Furthermore, the tracking success rate is also superior to previous methods, as IEKF successfully tracks the pose at all times, whereas the KF-less method losses a few frames and ArUco gives unstable tracking results. Also, as ArUco cannot track most frames in the dynamic landing dataset, results for that case in this scenario are omitted here.
	\begin{table}[ht]
		\centering
		\caption{\underline{Results from Relative State Estimation}}
		\label{tab:compare}
		\begin{tabular}{l|l|lll}
			&& IEKF     & KF-less \cite{faessler2014monocular}  & ArUco \cite{garrido2014automatic}      \\ 
			\hline\hline
			\multirow{3}{*}{Circle}    & ATE\_translation & 0.036 $m$  & 0.059 $m$  & 0.149 $m$    \\
									   & ATE\_rotation    & $3.104^\circ$ & $5.497^\circ$ & $41.77^\circ$  \\
									   & Success Rate             & 100 \%   & 99.7 \%  & 28.2 \%    \\
									   \hline
			\multirow{3}{*}{Dyn. Land} & ATE\_translation & 0.053 $m$  & 0.083 $m$  & N/A        \\
									   & ATE\_rotation    & $3.007^\circ$ & $8.956^\circ$ & N/A        \\
									   & Success Rate             & 100 \%   & 99.5 \%  & N/A       
			\end{tabular}
	\end{table}
	
	\subsection{Landing Experiment}
	\label{sec:EXP_LAND}
	Lastly, we present the landing mission. In specifics, the UGV is designed to move in two different motions, (1) linear motion and (2) circular motion. For dicussion, we plot the experiment position and velocity information, state estimation error, and the tracking error.
	
\begin{figure*}[b]
		\centering
		\includegraphics[width=1.0\textwidth]{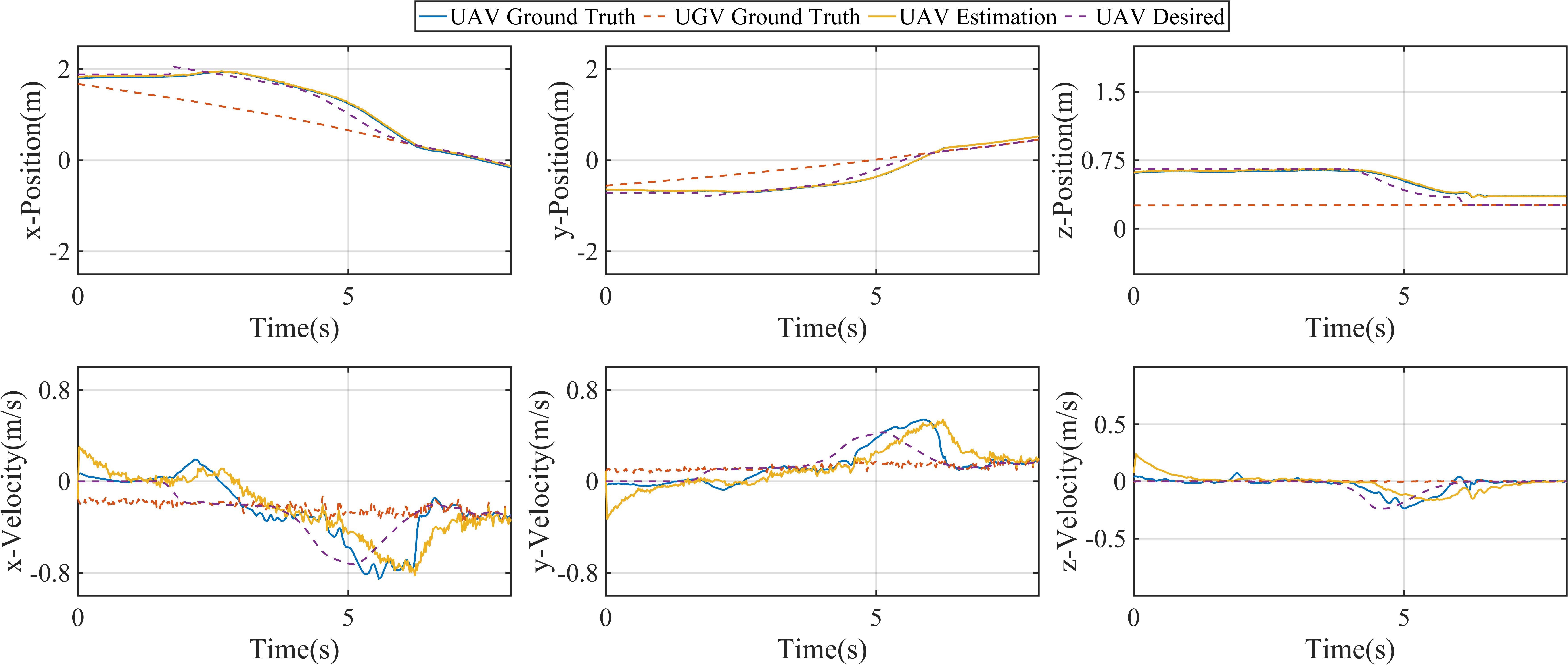}
		\caption{
			The position and velocity information of the linear landing mission. The maximum velocity of dynamic platform was 0.4 m/s, whereas the maximum velocity of the quadrotor was 0.9 m/s.
		}
		\label{fig:land_linear}
	\end{figure*}

	\textit{(1) Landing platform in linear motion:} 
    Figure \ref{fig:land_linear} encapsulates the linear landing experiment, plotting the position and velocity of both the landing platform and quadrotor. It could be seen that even without onboard exteroceptive sensors and computers, the quadrotor can perform landing on a constantly moving landing pad. We discuss more details in the circular case.

\begin{figure}[hbt!]		
		\centering
		\includegraphics[width=0.395\textwidth]{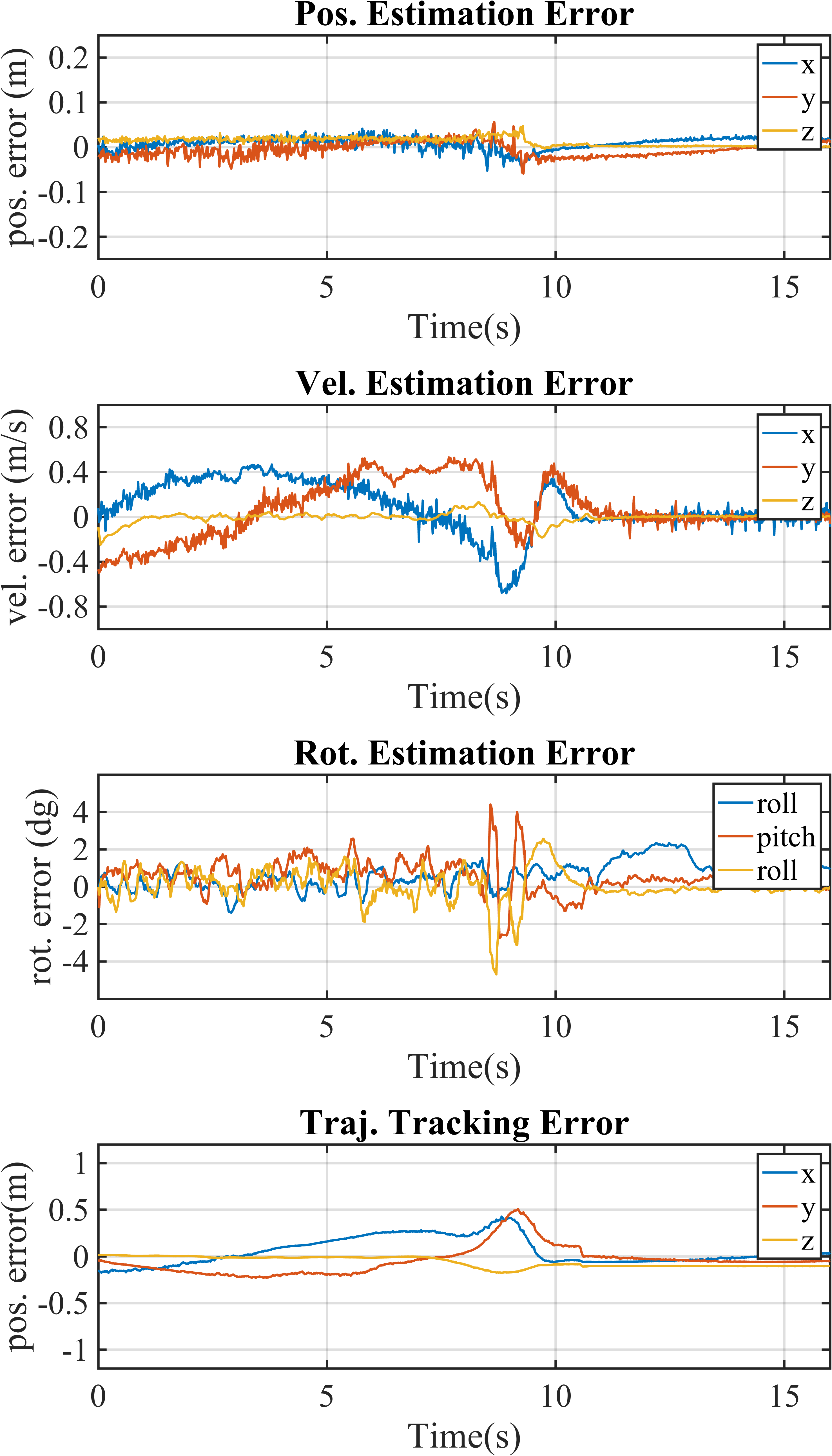}
		\caption{The estimation error and tracking error during the landing mission where the moving platform is in circular motion.}
		\label{fig:land_error}
		\vspace{-0.0cm}	 
	\end{figure}

	\textit{(2) Landing platform in circular motion:} 
	Similar to the previous experiment, when landing the quadrotor on a dynamic platform whose path is predefined to be a circle, the overall mission could be completed. In this case, we plot the position, rotation, and tracking error for further analysis. From figure \ref{fig:land_error}, the system gives stable estimation results regarding position during the landing. In contrast, the velocity error contains relatively huge differences and fluctuations. Nevertheless, without IMU data, we deem the results sufficiently accurate in this application. Furthermore, although high frequency noise could be found from rotation error, the overall results stay within the reasonable region.
	\begin{figure*}[t]
		\centering
		\includegraphics[width=1.0\textwidth]{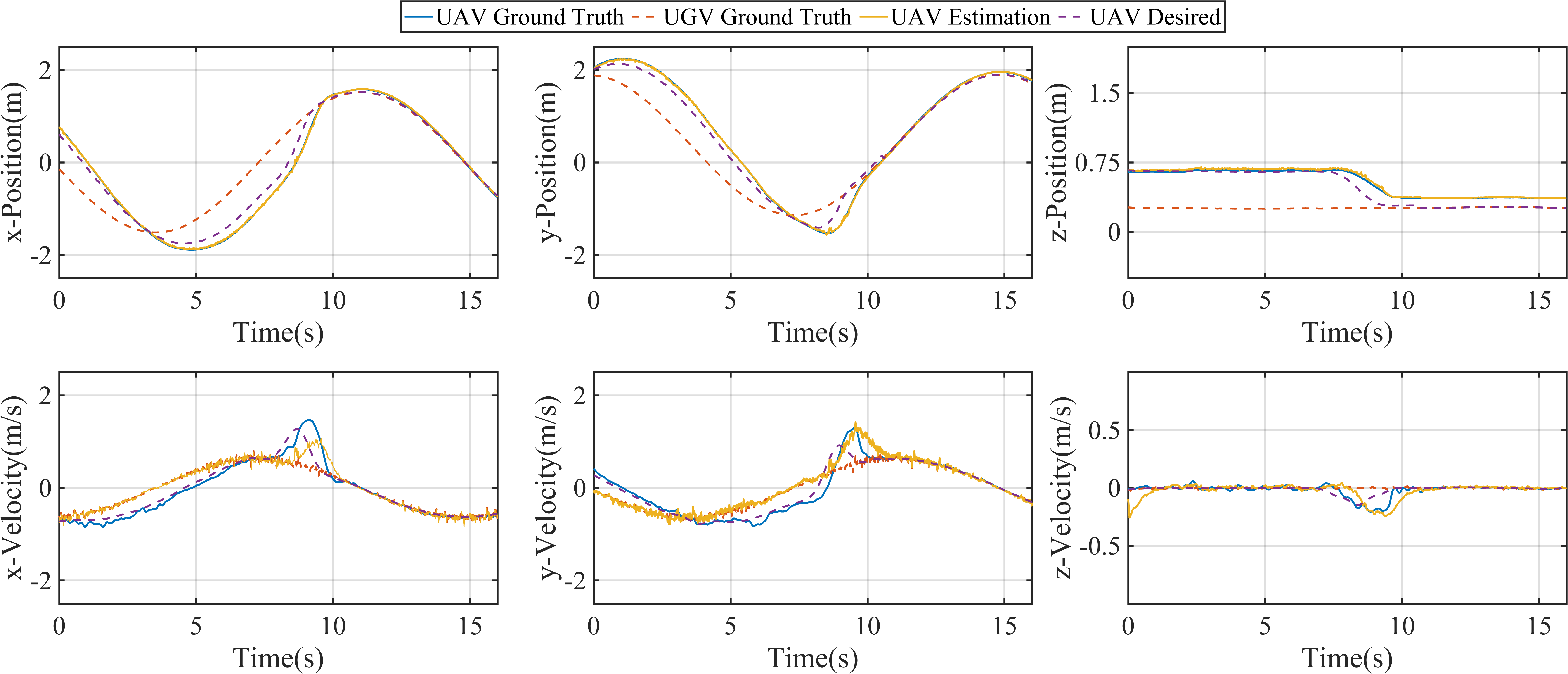}
		\caption{
			The position and velocity information of the circle landing mission. The maximum velocity of dynamic platform was 0.9 m/s, whereas the maximum velocity of the quadrotor was 1.8 m/s.
		}
		\label{fig:land_circle}
	\end{figure*}

	As for tracking error, the RMSE of tracking error is 0.2240 $m$. The culprits are the transmission latency and the lack of controller optimization. The former can be improved by adopting WiFi MESH configuration to remove dependencies on the central router. On the other hand, the latter can be improved by better PID tuning or adopting an advanced controller, which is out of the scope of the addressed work. Overall, we deem the experiments successful, and hence, the results imply the applicability of the proposed landing system. 

\section{Conclusion}
\label{sec:conclusion}
	In this study, we propose a novel quadrotor landing system, termed a ``non-robocentric'' (non-inertial) framework, designed to operate without onboard exteroceptive sensors and a computer. While we have validated the system in a controlled environment, we posit that this marks a crucial advancement toward an innovative autonomous landing framework, capable of mitigating the heavy reliance on onboard computers and exteroceptive sensors. This, in turn, has the potential to significantly reduce overall costs for the future large-scale deployment of UAV swarms in smart-city scenarios.

	In the near future, our research aims to extend the validation of our system by conducting experiments on outdoor moving vehicles and marine vessels, assessing its feasibility for real-world deployment. However, prior to these endeavors, we recognize the necessity for enhancements in the relative state estimation module, particularly focusing on improving range and stability. This aspect is acknowledged as part of our ongoing and future work.	

\bibliographystyle{IEEEtran}
\bibliography{references}

\begin{IEEEbiography}
[{\includegraphics[width=1in,height=1.25in,clip,keepaspectratio]{fig/lly}}]
	{Li-Yu Lo} received his Bachelor of Engineering degree from the Department of Aeronautical and Aviation Engineering at The Hong Kong Polytechnic University, Hong Kong, in 2021. He is currently a member in the AIRo-Lab, Department of Aeronautical and Aviation Engineering, The Hong Kong Polytechnic University. His research interests include sensor fusion, multi-robot pose estimation, and heterogeneous robotics systems.\\ \\
\end{IEEEbiography}

\begin{IEEEbiography}
[{\includegraphics[width=1in,height=1.25in,clip,keepaspectratio]{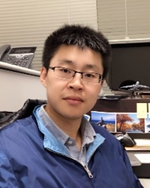}}]
	{Boyang LI} received his B.Eng. and M.Eng. degrees in Aeronautical Engineering from Northwestern Polytechnical University, Xi’an, China, in 2012 and 2015. He then received PhD degree in Mechanical Engineering from The Hong Kong Polytechnic University in 2019.

	He has conducted postdoctoral research at the Air Traffic Management Research Institute, Nanyang Technological University, Singapore and the School of Engineering, The University of Edinburgh, UK. In 2020, he established the Autonomous Aerial Systems Lab at The Hong Kong Polytechnic University. He joined the University of Newcastle as a lecturer in Aerospace Systems Engineering in 2023. His research interests include model predictive control, path/trajectory optimization, and field experiments of unmanned aircraft systems (UAS) and other mobile robots.
\end{IEEEbiography}

\begin{IEEEbiography}
[{\includegraphics[width=1in,height=1.25in,clip,keepaspectratio]{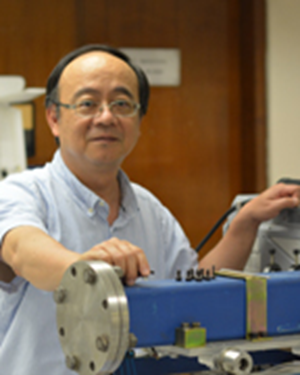}}]
	{Chih-Yung WEN} received his B.S. degree from the Department of Mechanical Engineering at National Taiwan University in 1986 and his M.S. and Ph.D. degrees from the Department of Aeronautics at the California Institute of Technology (Caltech) in 1989 and 1994, respectively.

	He joined the Department of Mechanical Engineering, The Hong Kong Polytechnic University in 2012, as a professor. In 2021, he became the chair professor and head of the Department of Aeronautical and Aviation Engineering in The Hong Kong Polytechnic University. His current research interests include modeling and control of tail-sitter UAVs, visual-inertial odometry systems for UAVs and AI object detection by UAVs.
\end{IEEEbiography}

\begin{IEEEbiography}
[{\includegraphics[width=1in,height=1.25in,clip,keepaspectratio]{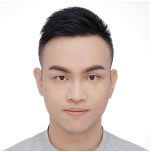}}]
	{Ching-Wei Chang} received his Bachelor of Engineering degree from the Department of Mechanical Engineering at Yuan Ze University, Taiwan, in 2015. He then received Ph.D. degree in Mechanical Engineering from The Hong Kong Polytechnic University in 2022. His research interests include VTOL UAVs, UAV inspection applications, and autonomous landing systems for UAVs. He currently holds the position of Researcher at the Hong Kong Center for Construction Robotics, affiliated with The Hong Kong University of Science and Technology.
\end{IEEEbiography}

\end{document}